\documentclass[12pt]{article}
\usepackage{amssymb,amsmath,cite,bm}
\usepackage[dvips]{graphicx,color}
\usepackage{psfrag,subfigure}
\begin{document}

\title{\bf Robust 3D Vision for Autonomous Robots}

\author{Farhad Aghili\thanks{email: faghili@encs.concordia.ca}}

\date{}

\maketitle

\begin{abstract}
This paper presents a fault-tolerant 3D vision system for autonomous robotic operation. In particular, pose estimation  of
space objects is achieved using 3D vision data in an integrated Kalman filter
(KF) and an Iterative Closest Point (ICP) algorithm in a closed-loop
configuration. The initial guess for the internal ICP iteration is
provided by state estimate propagation of the Kalman filer. The
Kalman filter is capable of not only estimating the target's states,
but also its inertial parameters. This allows the motion of the
target to be predictable as soon as the filter converges.
Consequently, the ICP can maintain pose tracking over a wider range
of velocity due to increased precision of ICP initialization.
Furthermore, incorporation of the target's dynamics model in the
estimation process allows the estimator continuously provide pose
estimation even when the sensor temporally loses its signal namely
due to obstruction. The capabilities of the pose estimation
methodology is demonstrated by a ground testbed for Automated
Rendezvous \& Docking (AR\&D). In this experiment, Neptec's Laser
Camera System (LCS) is used for real-time scanning of a satellite
model attached to a manipulator arm, which is driven by a simulator
according to orbital and attitude dynamics. The results showed that
robust tracking of the free-floating tumbling satellite can be
achieved only if the Kalman filter and ICP are in a closed-loop
configuration.
\end{abstract}


\section{Introduction}\label{Introduction}
There are several on-orbit servicing missions on the
horizon that will require capabilities for autonomous rendezvous \&
docking, autonomous rendezvous \& capture, proximity operation of
space vehicles, constellation satellites, and formation
flying~\cite{Aghili-Kuryllo-Okouneva-English-2010a,Aghili-2011k,zimpfer-spehar-1996,Aghili-2010f,Laliberte-Birglen-Gosselin-2002,yoshida-2003,Wingo-2004,Tatsch-Fitz-Coy-Gladun-2006,Aghili-Parsa-2009,Liang-Li-Xue-Qiang-2006,Aghili-Kuryllo-Okuneva-McTavish-2009,Kaiser-Rank-Krenn-2008,Aghili-2008c,Aghili-Parsa-2007b,Aghili-Parsa-Martin-2008a}.
Successful accomplishment of these missions critically relies on a
fault-tolerant sensor system  to obtain real-time relative pose
information (in 6-dof) of two spacecraft during proximity
operations. There are different vision systems capable of estimating
the pose (position and orientation) of moving objects. However,
among them, an active vision system such as the Neptec Laser Camera
System (LCS) is preferable because of its robustness in the face of
the harsh lighting conditions of
space~\cite{Samson-English-Deslauriers-Christie-2004,Aghili-Kuryllo-Okouneva-English-2010a}. As verified
during the STS-105 space mission, the 3D imaging technology used in
the LCS can indeed operate in space environment. The use of laser
range data has also been proposed for the motion estimation of
\textit{free-floating} space
objects~\cite{Linchter-Dubowsky-2004,Hillenbrand-Lampariello-2005}.
All vision systems, however, provide discrete and noisy pose data at
a relatively low rate, which is typically 1~Hz. It is worth
mentioning that another 3D sensor system called TriDAR is an
emerging technology that combines the complementary nature of LCS
triangulation and Lidar time-of-flight for enhancing the scan range
\cite{English-Zhu-Smith-Ruel-Christie-2005,Aghili-2016c}.

Vision-based pose estimation algorithms are 3D registration
processes, by which the range data set from different views are
aligned in a common coordinate system \cite{Kim-Hwang-2004,Aghili-2010s}. In
general, the 3D registration process consists of two steps: coarse
registration  and its refinement. The coarse registration process
gives a rough estimate of the 3D transformation by using a
positioning sensor that is then used as an initial 3D pose in the
refinement step. The ICP (Iterative Closest Point) algorithm is the
most widely used algorithm as the pose refinement method for
geometric alignment of slightly misaligned 3D
models~\cite{Besl-Mckay-1992,Greenspan-Yurick-2003}. Although the
basic ICP algorithm has proven to be very useful in the processing
of range data \cite{Greenspan-Yurick-2003}, a number of variations
on the basic method have been also developed to optimize different
phases of the algorithm
\cite{Greenspan-Yurick-2003,Rusinkiewicz-Levoy-2001,Aghili-Kuryllo-Okuneva-McTavish-2009,Godin-Laurendeau-Bergevin-2001,Chen-Medioni-1992}.

Taking advantage of the simple dynamics of a free-floating object,
which is not acted upon by any external force or moment, researchers
have employed different observers to track and predict the motion of
a target
satellite~\cite{Aghili-Kuryllo-Okouneva-English-2010c,Masutani-Iwatsu-Miyazaki-1994,Aghili-Parsa-2008b,Aghili-Parsa-2009}.
Using the Iterative Recursive Least Squares pose refinement
\cite{Drummond-Cipolla-2002}, the visual information from camera has
been used for relative pose estimation for autonomous rendezvous \&
docking \cite{Kelsey-Byrne-Cosgrove-Seereeram-Mehra-2006}. In some
circumstances, e.g., when there are occlusions, no sensor data is
available. Therefore, prediction of the motion of the target based
on the latest estimation of its states and dynamic parameters are
needed for a continuous pose estimation. This work is focused
on the integration of the Kalman filter and the ICP algorithm in a
closed-loop configuration for accurate and fault-tolerant pose
estimation of a free-falling tumbling satellite \cite{Aghili-Kuryllo-Okouneva-English-2010a}. In the conventional
pose estimation algorithm, the range data from a rangefinder scanner
along with a CAD-generated surface model are used by the ICP
algorithm to estimate the target's pose. The estimation can be made
more robust by placing the ICP algorithm and the KF estimator in a
closed-loop configuration, wherein the initial guess for the ICP is
provided by the estimate prediction \cite{Aghili-Salerno-2016}. The pose estimation based on
the ICP-KF configuration has the following advantages:
\begin{enumerate}
\item The KF reduced the measurement noise, thus providing a more
precise pose estimation.
\item Faster convergence of the ICP iteration and maintaining
tracking over a wider range of object velocity are achieved as a
consequence of increased precision of the ICP initial guess provided
by the KF.
\item Continuous pose estimation becomes possible even in the presence
of temporary sensor failure or obstruction as  a consequence of
 incorporation of the dynamic model of the target in the estimation process.
\end{enumerate}
The KF estimator is designed so that it can estimate not only the
target's states, but also its dynamic parameters. Specifically, the
dynamic parameters are the ratios of the moments of inertia of the
target, the location of its center of mass, and the orientation of
its principal axes. Not only does this allow long-term prediction of
the motion of the target, which is needed for motion planning, but
also it provides accurate pose feedback for the control system when
there are blackout, i.e., no observation data are available \cite{Aghili-Kuryllo-Okouneva-English-2010b,Aghili-Parsa-2008b}. In this
work, we use the Euler-Hill equations~\cite{Kaplan-1976} to derive a
discrete-time model that captures the evolution of the relative
translational motion of a tumbling target satellite with respect to
a chaser satellite which is freely falling in a nearby orbit.

\section{ICP Algorithm}
\label{sec:ICP}
This section reviews the basic ICP algorithm originally proposed by
Besl and McKay \cite{Besl-Mckay-1992}. The ICP algorithm is an
iterative procedure which minimizes a distance between point cloud
in one dataset and the closest points in the other. Suppose that we
are given a 3D point cloud dataset ${\cal U}=\{\bm u_1 \cdots \bm
u_m \}$ that corresponds to a single shape represented by {\em model
set} ${\cal M}$. It is known that for each point $\bm
u_i\in\mathbb{R}^3$ from the 3D points data set ${\cal U}$, there
exists at least one point, $\bm v_i$, on the surface of ${\cal M}$
which is closer to $\bm u_i$ than other points in ${\cal
M}$~\cite{Simon-Herbert-Kanade-1994}. The ICP algorithm assumes that
the initial rigid transformation $\{\bm A_0, \bm r_0 \}$ between
sets ${\cal U}$ and ${\cal M}$ is known;  $\bm r_0$ and $\bm A_0$
denote the translation vector and the rotation matrix, respectively.
Then, the problem of finding the correspondence between the two sets
can be formally expressed by
\begin{equation} \label{eq:ci}
\bm v_i = \mbox{arg} \min_{\bm v_j \in {\cal M} } \|\bm A_0 \bm
u_i + \bm r_0 - \bm v_j \| \qquad \forall i=1,\cdots,m  ,
\end{equation}
and then set ${\cal V} = \{\bm v_1 \cdots \bm v_m \}$ is formed
accordingly. Now, we have two independent sets of 3D points ${\cal
V}$ and ${\cal U}$ both of which corresponds to a single shape. The
problem is to find a fine alignment $\{ \bm A, \bm r \}$ which
minimizes the distance between these two sets of
points~\cite{Besl-Mckay-1992}. This can be formally stated as
\begin{align} \label{eq:ICP}
d=&  \mbox{arg} \min_{\bm r, \bm A} \frac{1}{m} \sum_{i=1}^m \| \bm
A \bm u_i + \bm r - \bm v_i \|^2 \qquad \forall \bm v_i \in {\cal
V}, \bm u_i \in {\cal U} \\ \notag & \text{s.t.} \quad  \bm A^T \bm
A = \bm 1_3,
\end{align}
where $\bm 1_3$ is the $3\times3$ identity matrix. There are several
closed-form solutions to the above least-squares minimization
problem \cite{Faugeras-Herbert-1986,Eggert-Lorusso-Fisher-1997}, of
which the quaternion-based algorithm \cite{Horn-1987} is the
preferable choice for $m>3$ over other methods
\cite{Besl-Mckay-1992}. The rotation matrix can be written as a
function of unit quaternion as
\begin{equation} \label{eq:R}
\bm A(\bm q) = (2q_o^2-1) \bm 1_3 + 2q_o [ \bm q_v \times ] + 2 \bm
q_v \bm q_v^T.
\end{equation}
where $\bm q_v$ and $q_o$ are the vector and scaler parts of the
quaternion, i.e., $\bm q=[\bm q_v^T \; q_o]^T$. Consider the
cross-covariance matrix of the sets ${\cal V}$ and  ${\cal U}$ given
by
\begin{equation}
\bm S = \mbox{cov}({\cal V}, {\cal U}) = \frac{1}{m} \sum_{i} \bm
v_i \bm u_i^T -  \bar{\bm v} \bar{\bm u}^T
\end{equation}
where
\begin{equation}
\bar{\bm v} = \frac{1}{m} \sum_{i=1}^m \bm v_i \quad \mbox{and}
\quad \bar{\bm u} = \frac{1}{m} \sum_{i=1}^m \bm u_i
\end{equation}
are the corresponding centroids of the sets. Then, it can be shown
that minimization problem \eqref{eq:ICP} is tantamount to find the
solution of the following quadratic programming \cite{Horn-1987}
\begin{equation} \label{eq:quadratic}
\max_{\| \bm q \| =1} \bm q^T \bm W \bm q
\end{equation}
where the symmetric weighting matrix $\bm W$ can be constructed from
cross-covariance matrix, $\bm S$, as follow
\begin{equation} \notag
\bm W =  \left[
\begin{array}{cccc}
s_{11}+s_{22} + s_{33} & \times & \times  & \times\\
s_{23}-s_{32} & s_{11}-s_{22}-s_{33} & \times  & \times\\
s_{31} -s_{13} & s_{21}+s_{12} &  -s_{11}+s_{22}-s_{33}  & \times\\
s_{12} -s_{21} & s_{31}+s_{13} &  s_{23}+s_{32} & -s_{11}-s_{22}+s_{33}
\end{array} \right]
\end{equation}
where $s_{ij}$ denotes the $ij$th entry of matrix $\bm S$. The unit
quaternion, $\breve{\bm q}$, that maximizes the quadratic function
of \eqref{eq:quadratic} is the eigenvector $\bm\xi_i$ corresponding
to the largest eigenvalue of the matrix $\bm W$, i.e., $\breve{\bm
q}=\bm\xi_i(\bm W)$ so that $\lambda_i(\bm W)=\lambda_{\rm max}$.
Then, the translation can be obtained from
\begin{equation}  \label{eq:r_bereve}
\breve{\bm r} = \bar{\bm v} - \bm A(\breve{\bm q}) \bar{\bm u}
\end{equation}
The ICP-based matching algorithm may proceed through the following
steps:
\begin{enumerate}
\item Given an initial coarse alignment $\{\bm q_0, \bm r_0 \}$, find the closest point pairs ${\cal V}$ from scan  3D points set ${\cal U}$ and model set ${\cal M}$
according to \eqref{eq:ci}. \label{it:CoarseAlign}
\item Calculate the fine alignment translation $\{\breve{\bm q} , \breve{\bm r} \}$ \label{it:FineAlign}
minimizing the mean square error the distance between two data sets
${\cal U}$ and  ${\cal V}$  according to
\eqref{eq:quadratic}-\eqref{eq:r_bereve}
\item Apply the incremental transformation from step
\ref{it:FineAlign} to step \ref{it:CoarseAlign}.
\item Iterate until either
the residual error $d$ is less than a threshold, $d_{\rm min}$, or
the maximum number of iteration is reached.
\end{enumerate}
It has been shown that the above ICP algorithm is guaranteed to
converge to a local minimum~\cite{Besl-Mckay-1992}. However, a
convergence to a global minimum depends on a good initial
alignment~\cite{Aghili-Salerno-2011,Amor-Ardabilian-Chen-2006}. In pose estimation of
moving objects, ``good'' initial poses should be provided at the
beginning of every ICP iteration. The initial guess for the pose can
be taken from the previous estimated pose obtained from the
ICP~\cite{Samson-English-Deslauriers-Christie-2004}. However, this
can make the estimation process fragile when dealing with relatively
fast moving target \cite{Aghili-Salerno-2010}. This is because, if the ICP does not converge
for a particular pose, e.g., due to occlusion, in the next
estimation step, the initial guess of the pose may be too far from
its actual value. If the initial pose happens to be outside the
global convergence region of the ICP process, from that point on,
the pose tracking is most likely lost for good. The estimation can
be made more robust by placing the ICP and a dynamic estimator in a
closed-loop configuration, whereby the initial guess for the ICP is
provided by the estimate prediction of the moving object. The
following sections describe the design of a Kalman filter which will
be capable of not only estimating the states but also the parameters
of a free-floating object.

\section{Motion Dynamics} \label{sec:modeling}

\begin{figure}[t]
\centering{\includegraphics[clip,width=10cm]{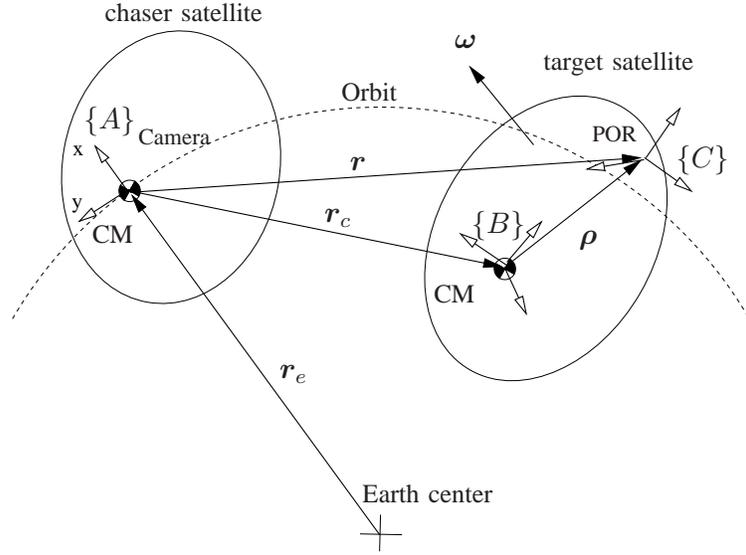}} \caption{The
body-diagram of chaser and target satellites moving in neighboring
orbits} \label{fig:chaisersat}
\end{figure}

Fig.~\ref{fig:chaisersat} illustrates the chaser and the target
satellites as rigid bodies moving in nearby orbits. Coordinate
frames $\{{\cal A}\}$ and $\{{\cal B}\}$ are attached to the chaser
and the target, respectively. The origins of frames $\{{\cal A}\}$
and $\{{\cal B}\}$ are located at center-of-masses (CM)s of the
observing and target satellites. The axes of $\{{\cal B}\}$ are
chosen to be parallel to the principal axes of the target satellite,
while $\{{\cal A}\}$ is orientated so that its $x$-axis is parallel
to a line connecting the Earth's center to the chaser CM and
pointing outward, and its $y$-axis lies on the orbital plane; see
Fig.~\ref{fig:chaisersat}. The coordinate frame $\{{\cal C}\}$ is
fixed to the target at its point of reference (POR) located at
distance $\bm\rho$ from the origin of $\{{\cal B}\}$. It is the pose
of $\{{\cal C}\}$ that is measured by the laser camera. We further
assume that the target satellite tumbles with angular velocity
$\bm\omega$. Also, notice that the coordinate frame $\{{\cal A}\}$
is not inertial; rather, it moves with the chaser satellite. Below,
vectors $\bm\rho$ and $\bm\omega$ are expressed in $\{{\cal B}\}$.

We choose to express the target orientation in the body-attached
frames coincident with its principal axes, $\{\cal B\}$, while the
translational motion variables are expressed in the body attached
frame of the chaser, $\{\cal A\}$. The advantages of such a mixed
coordinate selection are twofold: Firstly, the inertia matrix
becomes diagonal and will be independent of the target orientation,
thus resulting in a simple attitude model. Secondly, choosing to
express the relative translational motion dynamics in $\{\cal A\}$
leads to decoupled translational and rotational dynamics that
greatly contributes to the simplification of the mathematical model;
note that, since $\{\cal B\}$ is a rotating frame, the translational
acceleration seen from this frame depends on both the relative
angular velocity and relative acceleration. In the following,
subscripts $r$ and $t$ denote quantities associated with the
rotational and translational dynamics of the system, respectively.

\subsection{Review of Quaternion Kinematic}
In this section, we review some basic definitions and properties of
quaternions used in the rest of the paper. Consider quaternions $\bm
q_1$, $\bm q_2$, $\bm q_3$ and their corresponding rotation matrices
$\bm A_1$, $\bm A_2$, and $\bm A_3$, respectively. Operators
$\otimes$ and $\circledast$ are defined as
\begin{equation}\notag
\!\!\!\![\bm q\otimes]\triangleq\begin{bmatrix}-[\bm q_v \times] + q_o \bm 1_3 & \bm q_v\\
-\bm q_v^T&q_o \end{bmatrix}, \quad [\bm q \circledast] \triangleq
\begin{bmatrix} [\bm q_v \times] + q_o \bm 1_3 & \bm q_v\\ - \bm q_v^T  & q_o
\end{bmatrix}
\end{equation}
where $[\bm q_v \times]$ is the cross-product matrix of $\bm q_v$.
Then, $\bm q_3 =\bm q_1 \circledast \bm q_2 \equiv \bm q_2 \otimes
\bm q_1$ corresponds to product $\bm A_3=\bm A_1\bm A_2$. Also, the
conjugate quaternion \footnote{$q_o^*=q_o$ and $\bm q_v^*=-\bm
q_v$.} $\bm q^*$ of $\bm q$ is defined such that $\bm q^* \otimes
\bm q = \bm q \otimes \bm q^* = [0 \; 0 \; 0 \; 1 ]^T$. Moreover,
both $\otimes$ and $\circledast$ operations have associative
property, hence $\bm q_1 \otimes \bm q_2 \otimes \bm q_3$ and $\bm
q_1\circledast \bm q_2 \circledast \bm q_3$ being unambiguous \cite{Aghili-Salerno-2009}.

The orientation of $\{{\cal A}\}$ with respect to $\{{\cal C}\}$ is
represented by the unit quaternion $\bm q$, while the orientation of
$\{{\cal A}\}$ with respect to $\{{\cal B}\}$ is represented by the
unit quaternion $\bm\mu$. Moreover, $\bm\eta$ is a constant unit
quaternion which represent orientation of the principal axes, i.e.,
orientation of $\{{\cal B}\}$ with respect to $\{{\cal C}\}$.
Therefore, the following kinematic relation holds among the
aforementioned quaternions
\begin{equation} \label{eq:eta_mu}
\bm q = \bm\eta \otimes \bm\mu.
\end{equation}

\subsection{State Equations}

In the rest of this paper, the underlined form of any vector $\bm
a\in{\mathbb R}^3$ denotes the representation of that vector in
${\mathbb R}^4$, e.g., $\underline {\bm
  a}^T \triangleq [\bm a^T \; 0]$. Recall that the orientation of the principal axis of the target
satellite with respect to the chaser satellite is represented by a
unit quaternion $\bm\mu$.  The time derivative of $\bm\mu$ and the
relative angular velocity $\bm\omega_{\rm rel}$ are related by
\begin{equation} \label{eq:dotq}
\dot {\bm\mu} = \frac{1}{2} \underline {\bm\omega}_{\rm rel} \otimes
\bm\mu, \qquad \text{where} \qquad  \bm\omega_{\rm rel}=\bm\omega -
\bm\omega_n,
\end{equation}
and $\bm\omega_n$ is the angular velocity of the chaser satellite
expressed in the frame $\{{\cal B}\}$. Assume that the
attitude-control system of the chaser satellite makes it rotate with
the angular velocity of the reference orbit. Furthermore, denoting
the angular velocity of the reference orbit expressed in $\{{\cal
A}\}$ with
\begin{equation} \label{eq:Omega}
\bm n \equiv \begin{bmatrix} 0 & 0 & n \end{bmatrix}^T,
\end{equation}
one can relate $\bm n$ and $\bm\omega_n$ using the quaternion
transformation below:
\begin{equation} \label{eq:omeg_rel}
\underline{\bm\omega}_n = \bm\mu \otimes \underline {\bm n} \otimes
\bm\mu^*.
\end{equation}
Substituting \eqref{eq:omeg_rel} into \eqref{eq:dotq} and using the
properties of the quaternion products, we arrive at
\begin{align} \notag
\dot {\bm\mu} &= \frac{1}{2} \underline {\bm\omega} \otimes \bm\mu -
\frac{1}{2} (\bm\mu \otimes \underline {\bm n} ) \otimes (\bm\mu^*
\otimes \bm\mu) \\ \label{eq:dot_q_nonlin} &= \frac{1}{2} \big(
\underline {\bm\omega} \otimes - \underline {\bm n} \circledast
\big) \bm\mu
\end{align}
Define the variation of actual quaternion, $\bm\mu$, from its
nominal value, $\bar{\bm\mu}$, as
\begin{equation}\label{eq:deltaq-def}
\delta \bm\mu=\bm\mu\otimes\bar{\bm\mu}^*
\end{equation}
Then, adopting a technique similar to that used by other authors in
~\cite{Lefferts-Markley-Shuster-1982,Aghili-Parsa-2009}, we can
linearize the time derivative of the above equation about the
estimated states $\bar {\bm\mu}$ and $\bar{\bm\omega}$ to obtain
\begin{equation} \label{eq:delta_qv}
\frac{d}{dt} \delta \bm\mu_v \approx - \bar {\bm\omega} \times
\delta \bm\mu_v + \frac{1}{2} \delta \bm\omega.
\end{equation}

Assuming that the spacecraft is acted upon by three independent
disturbance torques about its principal axes and denoting the
principal moments of inertia as $I_{xx}$, $I_{yy}$, and $I_{zz}$, we
can write Euler's equations as
\begin{align} \notag
I_{xx} \dot \omega_x &= (I_{yy} - I_{zz}) \omega_y \omega_z +
\tau_x,\\ \notag I_{yy} \dot \omega_y &= (I_{zz} - I_{xx}) \omega_z
\omega_x + \tau_y,\\ \notag I_{zz} \dot \omega_z &= (I_{xx} -
I_{yy}) \omega_x \omega_y + \tau_z.
\end{align}
Rewriting the above equations in terms of the perturbation
$\epsilon_{\tau} =[\tau_x/I_{xx} \; \tau_y/I_{xx}
\;\tau_z/I_{xx}]^T$, we obtain
\begin{equation} \label{eq:dot_omega}
\dot {\bm\omega} = \bm\psi(\bm\omega) + \bm J(\bm p)
\bm\epsilon_{\tau},
\end{equation}
where $\bm p^T=[p_x\thickspace~p_y\thickspace~p_z]$, $p_x=(I_{yy}-
I_{zz})/I_{xx}$, $p_y=(I_{zz}- I_{xx})/I_{yy}$, and $p_z=(I_{xx}-
I_{yy})/I_{zz}$ are the inertia ratios, and
\begin{equation*}
\bm\psi(\bm\omega) = \begin{bmatrix} p_x \omega_y \omega_z \\ p_y \omega_x \omega_z \\
p_z \omega_x \omega_y  \end{bmatrix}, \quad  \bm J(\bm p) =
\mbox{diag} \Big\{1, \;\; \frac{1 - p_y}{1+p_x}, \;\;
\cfrac{1+p_z}{1-p_x} \Big\}.
\end{equation*}
Linearizing~\eqref{eq:dot_omega} about $\bar{\bm\omega}$ and
$\bar{\bm p}$ yields
\begin{equation} \label{eq:dot_omeg_param_lin}
\delta \dot{\bm\omega} = \bm M(\bar{\bm\omega}, \bar{\bm p})
\delta\bm\omega + \bm N(\bar{\bm\omega}) \delta \bm p + \bm
J(\bar{\bm p})\bm\epsilon_{\tau},
\end{equation}
\begin{align} \notag
\bm M(\bm\omega,\bm p) &= \frac{\partial\bm\psi}{\partial\bm\omega}
=
\begin{bmatrix} 0 & p_x \omega_z & p_x \omega_y \\ p_y \omega_z & 0 &
p_y \omega_x \\ p_z \omega_y & p_z \omega_x & 0
\end{bmatrix} \\ \notag
\bm N(\bm\omega) &= \frac{\partial \bm\psi}{\partial \bm p} =
\begin{bmatrix} \omega_y \omega_z & 0 & 0 \\ 0 & \omega_x \omega_z & 0 \\ 0 & 0 &
\omega_x \omega_y \end{bmatrix}.
\end{align}

The evolution of the relative position of the two satellites can be
described by the Euler-Hill model in {\em orbital mechanics}. Let
the chaser move on a circular orbit with radius $R_e$, thus the
carrier ray having an angular rate $n$. Further, assume that vector
$\bm r_c$ denotes the relative position of the CM's of the two
satellites expressed in the frame $\{{\cal A}\}$. Then, the
translational motion of the target satellite can be expressed as
\begin{equation} \label{eq:ddot_r}
\ddot {\bm r}_c = -2 \bm n \times \dot{\bm r}_c - \bm n \times(\bm n
\times \bm r_c) + \Big(- \mu_e \frac{\bm r_e + \bm r_c}{\| \bm r_e +
\bm r_c \|^3} + n^2 \bm r_e \Big) + \bm\epsilon_f,
\end{equation}
where $\mu_e$ is the gravitational parameter of the Earth and $\bm
r_e=[R_e \; 0 \; 0]^T$ is a constant position vector expressed in
frame $\{ {\cal A}\}$ that connects the Earth's center to the chaser
CM, and $\bm\epsilon_f$ denotes the perturbation of translational
motion. Note that the constant position vector can be computed from
the angular rate of the circular orbit by
\begin{equation} \label{eq:re}
\bm r_e= \begin{bmatrix} \sqrt[3]{\cfrac{\mu_e}{n^2}} & 0 & 0
\end{bmatrix}^T.
\end{equation}
The linearized equations of the translational motion, which are
known as the Euler-Hill equations \cite{Kaplan-1976} or the
Clohessy-Wiltshire equations \cite{Clohessy-Wiltshire-1960}, can be
written as
\begin{equation} \label{eq:translation_lin}
\delta \ddot{\bm r}_c = \bm K(n) \delta \bm r_c - 2 \bm n \times
\delta \dot{\bm r}_c + \bm\epsilon_f,
\end{equation}
where $\bm K(n) = \mbox{diag} \{ 3n^2 \; 0 \; -n^2 \}$.

In addition to the inertia of the target satellite, the location of
CM, $\bm\rho$, and the orientation of the principal axis denoted by
unit quaternion,  $\bm\eta$, are uncertain.  Now assume that the
state vector to be estimated is:
\begin{equation}
\bm x=[\bm\mu_v^T \; \bm\omega^T \; \bm p^T \; \bm r_c^T \; \dot{\bm
r}_c^T \; \bm\rho^T \; \bm\eta_v^T]^T ~\in \mathbb{R}^{21}
\end{equation}
where $\bm\eta_v$ is the vector part of unit quaternion $\bm\eta$.
We assume that the dynamics parameters of the target remain
constant, i.e.,
\begin{equation} \label{eq:dynparam}
\dot{\bm p} = \dot{\bm\rho} = \dot{\bm\eta_v} = \bm 0_{1\times 3}
\end{equation}
One can combine the nonlinear equations \eqref{eq:dot_q_nonlin},
\eqref{eq:dot_omega}, \eqref{eq:ddot_r} and \eqref{eq:dynparam} in
the standard form as $\dot{\bm x} = \bm f( \bm x, \bm\epsilon)$,
which can be used for state propagation. However, the KF design also
requires the state-transition matrix of the linearized system. Let
us define the variation of quaternion $\bm\eta$ with respect to its
nominal value $\bar{\bm\eta}$ according to
\[ \delta \bm\eta_=\bar{\bm\eta}^* \otimes \bm\eta. \]
Then setting the linearized equations  \eqref{eq:delta_qv},
\eqref{eq:dot_omeg_param_lin}, and \eqref{eq:translation_lin} in the
standard state-space form yields:
\begin{equation} \label{eq:continous}
\delta \dot{\bm x} = \bm F \delta {\bm x} + \bm B \bm \epsilon
\end{equation}
where vector $\bm\epsilon=[\bm\epsilon_{\tau}^T \;
\bm\epsilon_f^T]^T$ contains the entire process noise, and
\begin{equation} \notag
\bm F = \begin{bmatrix} - [\bar{\bm\omega} \times] & \frac{1}{2} \bm
1_3 & \bm 0_3 & \bm 0_3 & \bm 0_3 & \bm 0_{3\times6} \\
\bm 0_3 & \bm M(\bar{\bm\omega}, \bar{\bm p}) & \bm
N(\bar{\bm\omega}) & \bm 0_3 & \bm 0_3 & \bm 0_{3\times6} \\
\bm 0_3 & \bm 0_3 & \bm 0_3 & \bm 0_3 & \bm 0_3 & \bm 0_{3\times6} \\
\bm 0_3 & \bm 0_3 & \bm 0_3 & \bm 0_3 & \bm 1_3 & \bm 0_{3\times6}  \\
\bm 0_3 & \bm 0_3 & \bm 0_3 & \bm K(n) & -2 [\bm n \times] & \bm 0_{3\times6} \\
\bm 0_{6\times 3} & \bm 0_{6\times 3} & \bm 0_{6\times 3} & \bm
0_{6\times 3} & \bm 0_{6\times 3} & \bm 0_6
\end{bmatrix},
\end{equation}
\begin{equation} \notag
\bm B = \begin{bmatrix} \bm 0_3 & \bm 0_3 \\
\bm J(\bar{\bm p}) & \bm 0_3 \\
\bm 0_{6 \times 3} & \bm 0_{6 \times 3} \\
\bm 0_3 & \bm 1_3 \\
\bm 0_{6 \times 3} & \bm 0_{6 \times 3}
\end{bmatrix}.
\end{equation}

The equivalent discrete-time system  \eqref{eq:continous} can be
written as
\begin{equation} \label{eq:discere_sys}
\delta \bm x_{k+1} = \bm\Phi_k \delta \bm x_k + \bm w_k.
\end{equation}
Here the solution to the state transition matrix $\bm\Phi_k$ and
discrete-time process noise $\bm Q_k=E[\bm w_k \bm w_k^T]$ can be
obtained based on the van Loan method \cite{vanLoan-1978} as
$\bm\Phi_k=\bm D_{22}^T$ and $\bm Q_k=\bm\Phi_k \bm D_{12}$, where
\[ \bm D = \begin{bmatrix} \bm D_{11} & \bm D_{12} \\ \bm 0 & \bm D_{22} \end{bmatrix} = \exp \Big( \begin{bmatrix} -\bm F &
\bm B \bm\Sigma \bm B^T \\ \bm 0 & \bm F^T
\end{bmatrix} T \Big)\] with $T$ being the sampling time and
$\bm\Sigma = E[ \bm\epsilon \bm\epsilon^T]=\mbox{diag}
\{\sigma_{\tau}^2 \bm 1_3, \sigma_{f}^2 \bm 1_3 \}$.

\section{Observation Equations}

The ICP algorithm gives noisy measurements of the position,
$\breve{\bm r}$, and orientation, $\breve{\bm q}$, of the target. In
the followings, the measurement variables will be expressed in terms
of the states variables. It is apparent from \eqref{fig:chaisersat} that
\begin{equation} \label{eq:r}
\bm r = \bm r_c  + \bm A(\bm\mu) \bm\rho,
\end{equation}
Also, recall from  \eqref{eq:eta_mu} that quaternion $\bm q$ can be
described as a function of $\bm\eta$ and $\bm\mu$, i.e., $\bm
q(\bm\mu,\bm\eta) = \bm\eta \otimes \bm\mu$. Consider nominal
quaternions $\bar{\bm\eta}$ and $\bar{\bm\mu}$ which are chosen to
be close to the actual quaternions; as will be later discussed in
the next section. Then, the variation of the quaternion, $\delta \bm
q$, with respect to nominal quaternions is defined as
\begin{equation}\label{eq:mu-def}
\delta \bm q = \bar{\bm\eta}^*  \otimes \bm q \otimes \bar{\bm\mu}^*
= \delta \bm\eta \otimes \delta \bm\mu.
\end{equation}
Thus, the observation equation can be written as
\begin{equation}
\bm z = \bm h(\bm x) + \bm\nu
\end{equation}
where $\bm\nu$ is the measurement noise, which is assumed to be
white with the covariance $E[\bm\nu \bm\nu^T]=\bm R$, and
\begin{equation} \label{eq:h_nonlin}
\bm z = \begin{bmatrix} \breve{\bm r} \\ \delta \breve{\bm q}
\end{bmatrix}, \qquad \bm h(\bm x) = \begin{bmatrix} \bm r_c  + \bm A(\delta \bm\mu \otimes \bm\mu)
\bm\rho \\ \mbox{vec}(\delta \bm\eta \otimes \delta \bm\eta)
\end{bmatrix}.
\end{equation}

Note that the observation vector \eqref{eq:h_nonlin} is a nonlinear function of the states\cite{Aghili-2010p}. To
linearize the observation vector, one also needs to derive the
sensitivity of the nonlinear observation vector with respect to the
system states. It can readily be seen from~\eqref{eq:R} that, for a small rotation
$\delta\bm\mu$, i.e., $\|\delta \bm\mu_v\| \ll 1$ and $\delta
\mu_0\approx 1$, the first-order approximation of the rotation
matrix can be written as
\[\bm A(\delta \bm\mu) \approx \bm 1_3 + 2 [\delta \bm\mu_v \times] \]
Hence, \eqref{eq:r} can be approximated as
\begin{equation} \label{eq:r_approx}
\bm r \approx \bm r_c + \bar{\bm A} (\bm\rho + 2 \delta \bm\mu_v
\times \bm\rho),
\end{equation}
where $\bar{\bm A}= \bm A(\bar{\bm\mu})$. Moreover, in view of the
definition of the quaternion product and assuming small quaternion
variations, we can say that
\begin{equation} \label{eq:qv_approx}
\delta \bm q_v \approx \delta \bm\mu_v \times \delta \bm\eta_v +
\delta \bm\mu_v + \delta \bm\eta_v.
\end{equation}
It is apparent from \eqref{eq:r_approx} and \eqref{eq:qv_approx}
that the measurement equations are bilinear functions of the states.
Consequently, we can derive the sensitivity matrix $\bm H
=\frac{\partial \bm h}{\partial \bm x}$ as
\begin{align*}
\bm H  = \left[\begin{array}{ccccccc} -2\bar{\bm A} [\bm\rho
\times] &\bm 0_3 &\bm 0_3 &
\bm 1_3 & \bm 0_{3}   & \bar{\bm A}(\bm 1_3+ 2[\delta \bm\mu_v \times]) &\bm 0_{3}  \\
-[\delta \bm\eta_v \times] + \bm 1_3 &\bm 0_3 &\bm 0_3 & \bm 0_{3}
& \bm 0_{3}  & \bm 0_{3}& [\delta \bm\mu_v \times] + \bm 1_3
\end{array} \right]
\end{align*}

\subsection{Filter design}

Recall that $\delta \bm\mu_v$ is a small deviation from the the
nominal trajectory $\bar{\bm\mu}$. Since the nominal angular
velocity $\bar{\bm\omega}_k$ is assumed constant during each
interval, the trajectory of the nominal quaternion can be obtained
from
\[  \bar{\bm\mu}_k = e^{\frac{1}{2} T [\hat{\underline{\bm\omega}}_{k-1} \otimes}] \hat {\bm\mu}_{k-1}. \]
However, since $\bm\eta$ is a constant variable, we can say $\bar
{\bm\eta}_k = \hat{\bm\eta}_{k-1}$. Let us assume that the state
vector is partitioned as $\bm x =[\bm\mu_v^T \; \bm\chi^T \;
\bm\eta_v^T]^T$. Then, the EKF-based observer for the associated
noisy discrete system \eqref{eq:discere_sys} is given in two steps:
($i$) estimate correction
\begin{subequations}
\begin{align} \label{eq:K_est}
\bm K_k & = \bm P_k^- \bm H_k^T \big(\bm H_k \bm P_k^- \bm H_k^T+ \bm R_k \big)^{-1} \\
\label{eq:x_est}
\begin{bmatrix} \delta\hat{\bm\mu}_{v_k} \\ \hat{\bm\chi}_k \\
\delta\hat{\bm\eta}_{v_k} \end{bmatrix} &= \begin{bmatrix} \delta\hat{\bm\mu}_{v_k}^- \\ \hat{\bm\chi}_k^- \\
\delta\hat{\bm\eta}_{v_k}^- \end{bmatrix} + \bm K \big(\bm z_k - \bm
h(\hat{\bm x}_k^-) \big) \\ \label{eq:P_est} \bm P_k &= \big( \bm
1_{21} - \bm K_k \bm H_k \big) \bm P_k^-
\end{align}
\end{subequations}
and ($ii$) estimate propagation
\begin{subequations}
\begin{align}\label{eq:state-prop}
\hat{\bm x}_{k+1}^- &=\hat{\bm x}_k + \int_{t_k}^{t_{k+1}} \bm f(\bm x(t),0)\,{\text d}t\\
\bm P_{k+1}^- &= \bm\Phi_k \bm P_k \bm\Phi_k^T + \bm Q_k
\end{align}
\end{subequations}
The unit quaternions are computed right after the innovation step
\eqref{eq:x_est} from
\begin{align*}
\hat{\bm\mu}_k  &=
\begin{bmatrix} \delta \hat{\bm\mu}_{v_k} \\ \big( 1- \| \delta \hat{\bm\mu}_{v_k} \|^2 \big)^{1/2}
\end{bmatrix} \otimes e^{\frac{1}{2} T [
\hat{\underline{\bm\omega}}_{k-1} \otimes]} \hat{\bm\mu}_{k-1},\\
\hat{\bm\eta}_k & =
\begin{bmatrix} \delta \hat{\bm\eta}_{v_k} \\ \big(1 - \|  \hat{\bm\eta}_{v_k} \|^2
\big)^{1/2} \end{bmatrix} \odot \hat{\bm\eta}_{k-1}.
\end{align*}

\begin{figure}
\begin{center}
\subfigure[Each ICP iteration uses the last iteration's pose
estimate to seed the next iteration.]
{\includegraphics[clip,width=8.5cm]{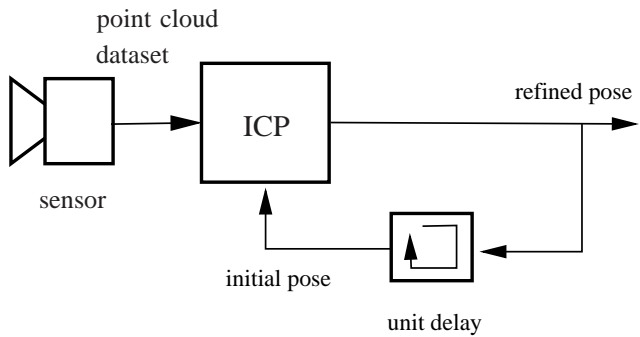} \label{fig:delay_icp} }
\subfigure[Each ICP iteration uses the KF pose prediction for the
time a new dataset is acquired by the sensor.]
{\includegraphics[clip,width=10cm]{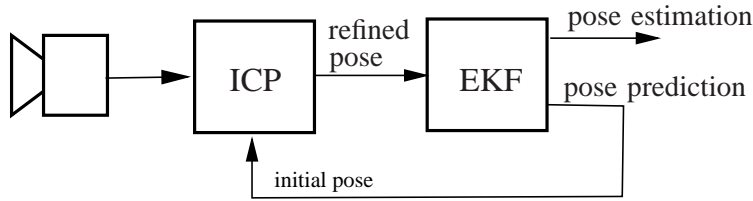} \label{fig:ekf_icp}}
 \caption{Pose tracking of moving objects.}
\label{fig:ekf_lcs}
\end{center}
\end{figure}

\subsection{ICP-KF Closed-Loop Configuration}
The ICP basically is a search algorithm which tries to find the best
possible match between the 3D data of the LCS and a model within the
neighborhood of the previous pose. In other words, the LCS
sequentially estimates the current pose based on the previous one.
That makes the estimation process fragile. This is because if the
ICP does not converge for a particular pose, then, in the next
estimation step, the initial guess of the pose may be far from the
actual one. If the initial pose occurs outside the convergence
region of the ICP process, from that point on the pose tracking is
lost for good. Fig.~\ref{fig:delay_icp} illustrates the conventional
ICP loop where each iteration uses the last iteration's pose
estimate to seed the next iteration either for a fixed number of
iterations, fixed time limit, or some convergence criteria are met.
The refined pose obtained from the ICP algorithm is then considered
as the pose estimation. In this configuration, a new dataset is
acquired and the output pose from the last dataset is normally used
as the initial pose for the new dataset.

On the other hand, Fig.~\ref{fig:ekf_icp} illustrates the ICP and KF
in a closed-loop configuration where the refined pose is used as a
input to KF while, in turn, the pose prediction of KF is used as the
initial pose for the ICP. More specifically, the latter pose
estimation process takes the following steps:
\begin{enumerate}
\item Refine an initial given pose by applying the ICP iteration on
a dataset acquired by the laser sensor. \label{item:refine}
\item Input the refined pose to the KF in order to filter out the sensor noise and
to estimate the object velocities as well as its dynamic parameters.
\item Use the estimation of the states and parameters to propagate the
pose at the time when a new dataset is acquired by the sensor.
\item Go to step \ref{item:refine} and use the pose prediction as the initial
pose.
\end{enumerate}

\section{Experiment}

\begin{figure}
\centering\subfigure[The satellite CAD
model.]{\includegraphics[clip,width=9cm]{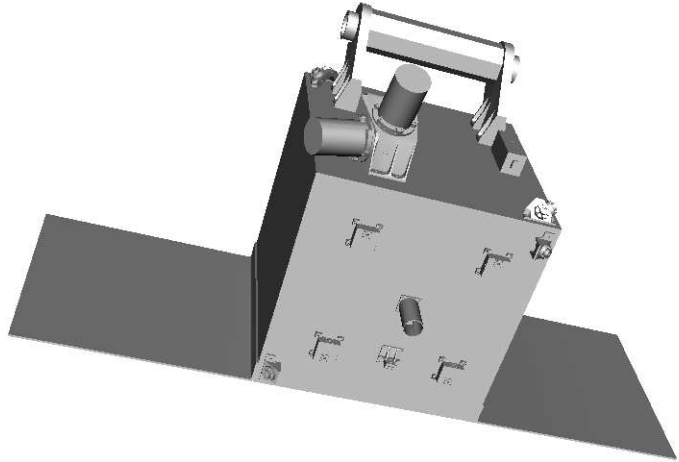}\label{fig:cad}}
\centering\subfigure[The point-cloud data from scanning of the
satellite.]{\includegraphics[clip,width=10cm]{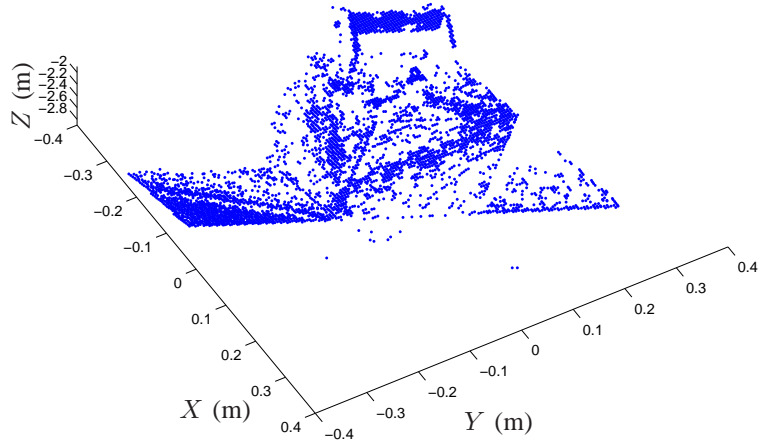}\label{fig:pointcloud}}
\caption{Matching points from the CAD model and the scanned data to
estimate the satellite pose.} \label{fig:cad_icp}
\end{figure}

In this section, experimental results are presented that  show
comparatively the performance of the pose estimation with and
without KF in the loop. The
experimental setup includes a satellite mockup  attached to a
manipulator arm, which is driven by a simulator according to orbital
dynamics~\cite{Aghili-Kuryllo-Okouneva-English-2010a}. The simulator generates the evolution of the relative
position and orientation of the two satellites  according to
\eqref{eq:dot_q_nonlin} and \eqref{eq:ddot_r}. The Neptec laser
rangefinder scanner is used
to obtain the pose measurements at a rate of 2~Hz. For the
spacecraft simulator that drives the manipulator, parameters are
selected as
\begin{equation} \notag
\bm I_c= \begin{bmatrix}4 & 0 & 0 \\
0& 8 & 0 \\ 0& 0 & 5 \end{bmatrix}~(\mbox{kgm}^2) \quad \mbox{and}
\quad \bm\rho=\begin{bmatrix} -0.15 \\ 0 \\ 0
\end{bmatrix}~(\mbox{m}).
\end{equation}

The solid model of the satellite mockup, Fig.~\ref{fig:cad}, and the
point-cloud data, generated by the laser rangefinder sensor,
Fig.~\ref{fig:pointcloud}, are used by the ICP algorithm to estimate
the satellite's pose according to the ICP initialization
configurations illustrated in Fig.~\ref{fig:ekf_lcs}. Three test
runs conduced to demonstrate the capability of the estimator are:
(i) identification of the spacecraft dynamics parameters, (ii)
accuracy and robustness of pose tracking at high velocity.

\begin{figure}
\centering{\includegraphics[clip,width=12cm]{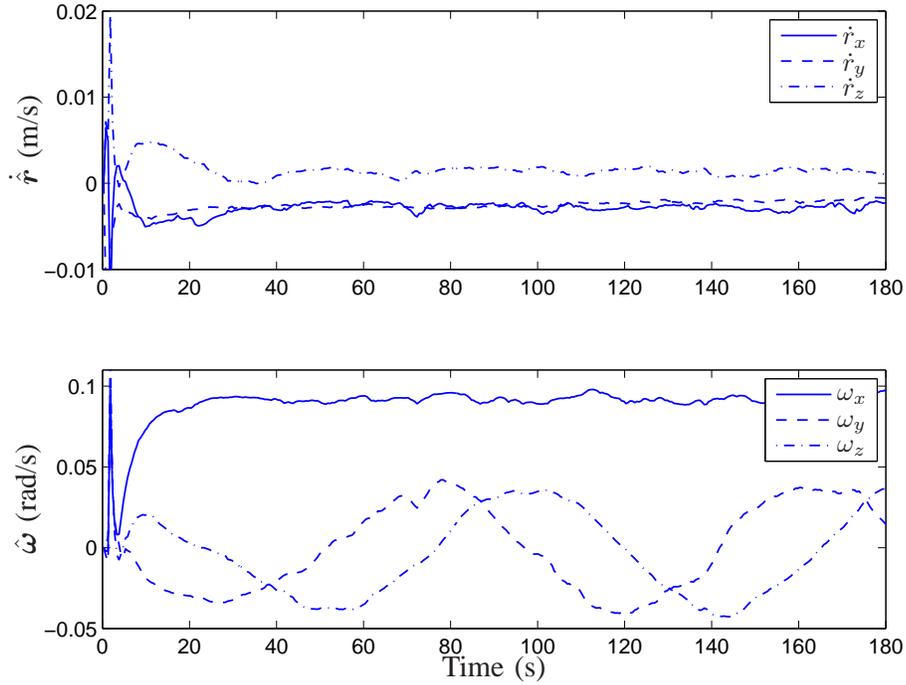}
\caption{Estimations of the satellite's linear and angular
velocities}} \label{fig:velocity}
\end{figure}

\begin{figure}
\centering{\includegraphics[clip,width=12cm]{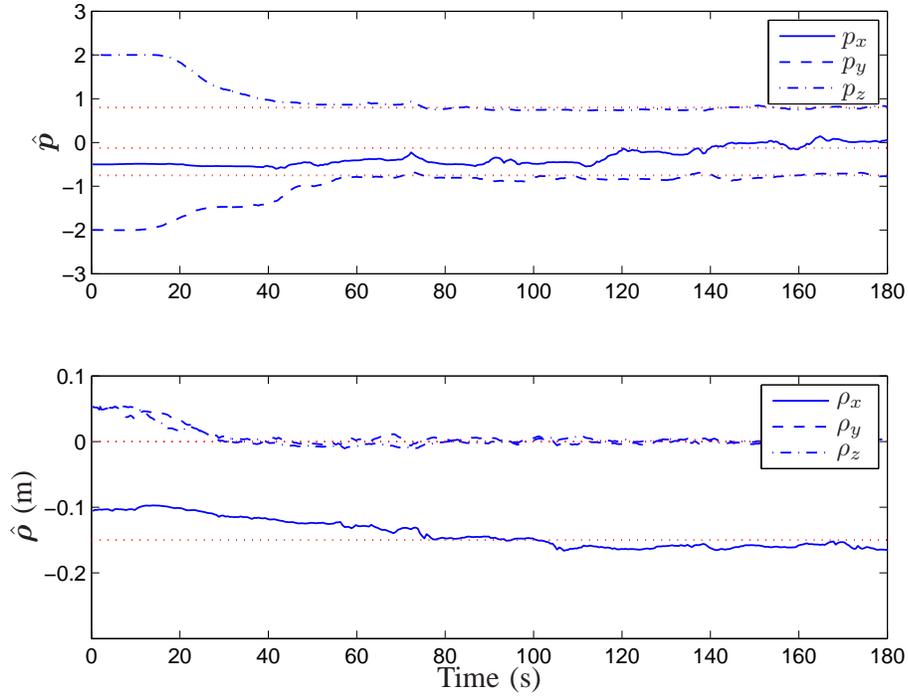}
\caption{Estimations of the satellite's dynamics parameters versus
their actual values depicted by dotted lines.}} \label{fig:paramters}
\end{figure}

\begin{figure}
\centering{\includegraphics[clip,width=12cm]{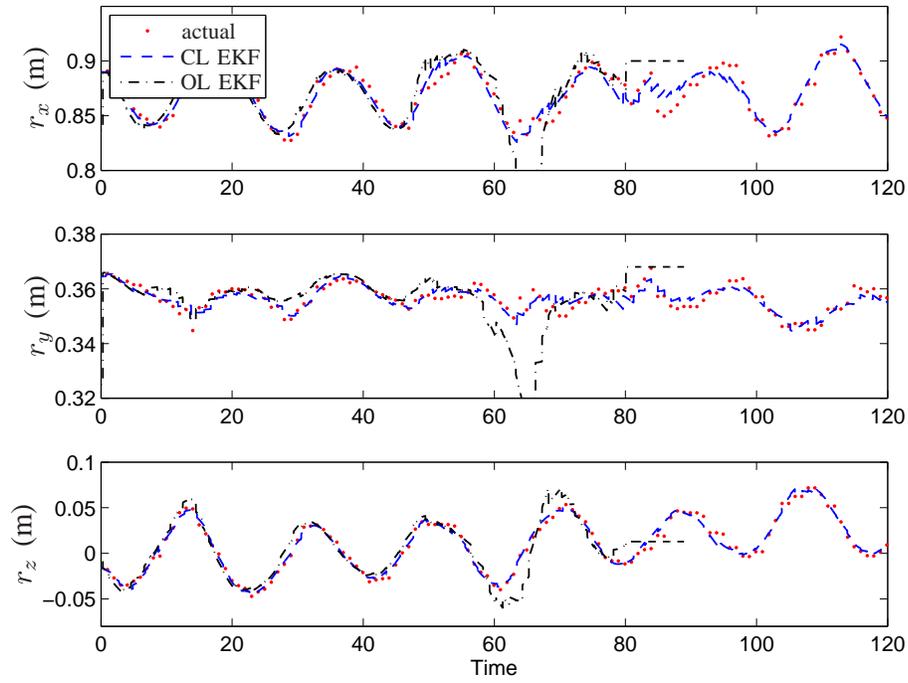}
\caption{Tracking of the satellite's position.}}
\label{fig:position_track}
\end{figure}

\begin{figure}
\centering{\includegraphics[clip,width=12cm]{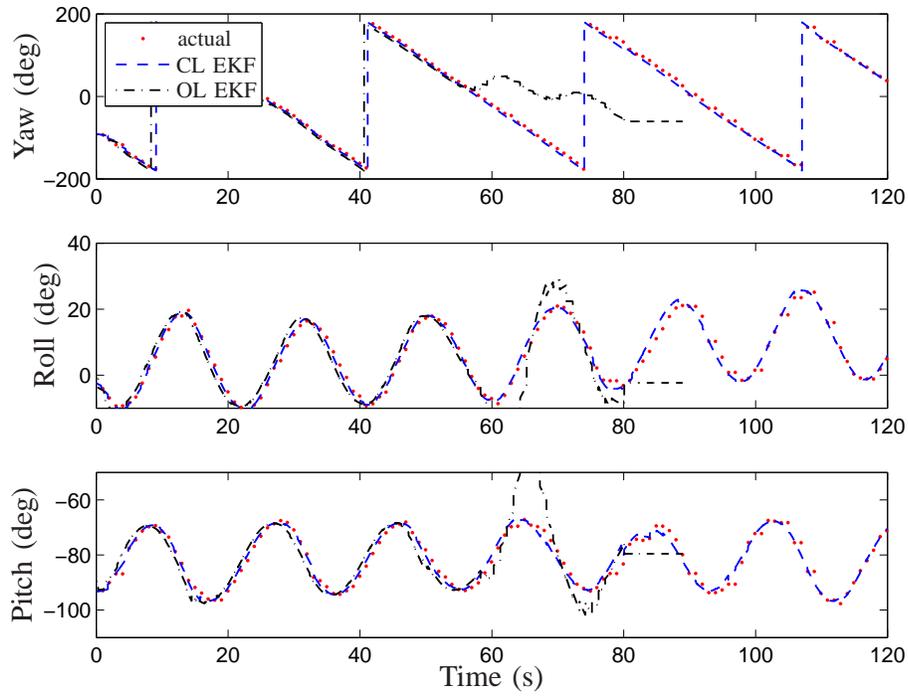}
\caption{Tracking of the satellite's attitude.}}
\label{fig:attiude_track}
\end{figure}

\begin{figure}
\centering{\includegraphics[clip,width=12cm]{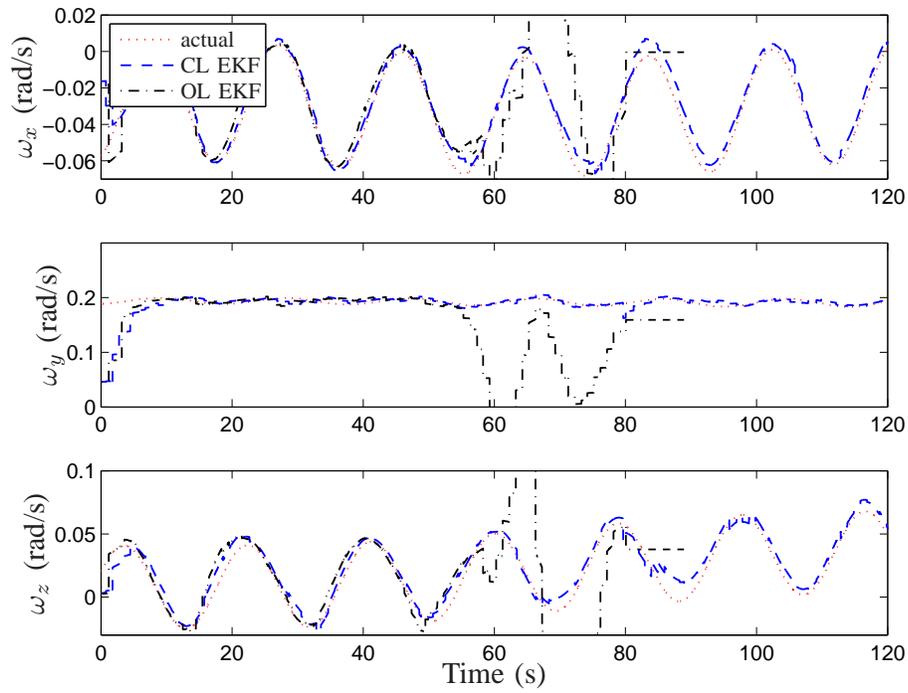}
\caption{Satellite's angular velocities.}} \label{fig:omega_track}
\end{figure}

\begin{figure}
\centering{\includegraphics[clip,width=12cm]{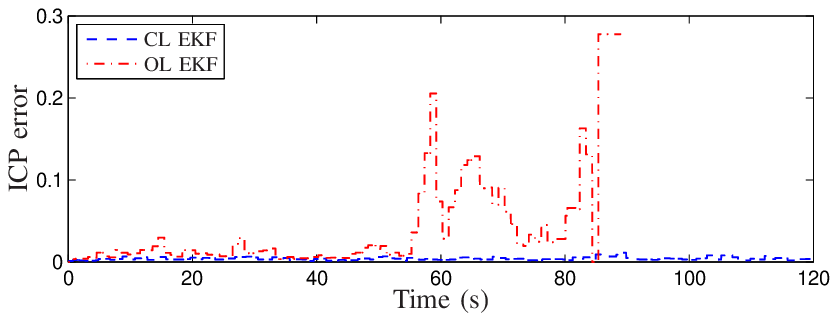}
\caption{Normalized ICP fit metric.}} \label{fig:icp_track}
\end{figure}

\subsection{Identification of Dynamics Parameters}
Figs.~\ref{fig:velocity} and \ref{fig:paramters} show the
trajectories of the estimated velocities and parameters,
respectively. The true values of the parameters are depicted by
dotted lines in Fig.~\ref{fig:paramters}. It is evident from the
graphs that the estimated dynamics parameters converged to their
true values after about 120~s. The slow convergence rate of the
estimator is attributed to the rather small angular velocity of the
target satellite. As will be later shown in the following section,
the convergence time improves at a higher target velocity.

The robustness of the pose estimation of the tumbling satellite with
and without incorporating the KF are comparatively illustrated in
Figs~\ref{fig:position_track}, \ref{fig:attiude_track}, and
\ref{fig:omega_track}. Note that here the quaternion is converted to
the the Euler's angles for representing the orientation. The figures
compare the actual trajectories of satellite pose and its angular
velocity, obtained by making use of the manipulator kinematic model
and measurements of the joint angles and velocities, with those
estimated from the vision system. In order to illustrate the
importance of the ICP and KF closed-loop configuration in a robust
pose estimation, the motion of the target satellites are estimated
by two methods: i) Using ICP and EKF in a closed configuration as
shown in Fig.~\ref{fig:ekf_icp}, and ii) cascaded ICP and EKF in an
open loop configuration, i.e., the KF does not provide the ICP with
the initial guess as shown in Fig.~\ref{fig:delay_icp}. The
associated trajectories of the former and the later methods are
labeled by CL EKF and OL EKF in the figures, respectively. The
motion estimation trajectories are compared versus the actual
trajectories calculated from the values of the manipulators' joint
angle and velocity measurements by making use of the robot
kinematics model. It is evident from the figures that the ICP-based
pose estimation is fragile if the initial guess is taken from the
last estimated pose. This will cause a growing ICP fit metric over
time, as shown in Fig.~\ref{fig:icp_track}, that eventually leads to
a total failure at $t=55$~sec. On the other hand, the pose
estimation with the ICP and the KF in the closed-loop configuration
exhibits robustness. Trajectories of the estimated angular
velocities obtained from the KF versus the actual trajectories are
illustrated in Fig.~\ref{fig:omega_track}. The plot shows that the
estimator converges at around $t=8$~s.

\section{Conclusion}
A robust 3D vision system for autonomous operation of space robots has been presented. 
A method for pose estimation of free-floating space objects which
incorporates a dynamic estimator in the ICP algorithm has been
developed. A Kalman filter was used for estimating the relative pose
of two free-falling satellites that move in close orbits near each
other using position and orientation data provided by a laser vision
system.  Not only does the filter estimate the system states, but
also all the dynamics parameters of the target. Experimental results
obtained from scanning a tumbling satellite mockup demonstrated that
the pose tracking based on the ICP alone did not
converge. On the other hand, the integration scheme of the KF and
ICP yielded a robust pose tracking.

\bibliographystyle{IEEEtran}


\end{document}